\documentclass[peerreview]{IEEEtran}
\usepackage{cite} 
\usepackage{url} 
\usepackage[utf8]{inputenc} 
\usepackage{booktabs} 
\usepackage{graphicx}
\usepackage{graphicx,amssymb,amstext,amsmath,eqnarray,lipsum,mathtools}
\usepackage{bm} 
\usepackage{hyperref}

\DeclarePairedDelimiter\floor{\lfloor}{\rfloor}

\begin{document}

\title{Uncertainty Propagation in Convolutional Neural Networks \\ --Technical Report--}

\author{Christos Tzelepis and Ioannis Patras \\
School of Electronic Engineering and Computer Science\\
Queen Mary, University of London\\
}

\maketitle
\tableofcontents
\listoffigures

\IEEEpeerreviewmaketitle
\begin{abstract}
In this technical report we study the problem of propagation of uncertainty (in terms of variances of given uni-variate normal random variables) through typical building blocks of a Convolutional Neural Network (CNN). These include layers that perform linear operations, such as 2D convolutions, fully-connected, and average pooling layers, as well as layers that act non-linearly on their input, such as the Rectified Linear Unit (ReLU). Finally, we discuss the sigmoid function, for which we give approximations of its first- and second-order moments, as well as the binary cross-entropy loss function, for which we approximate its expected value under normal random inputs. A PyTorch implementation of the presented ``uncertainty-aware'' layers is available under the MIT license here: \href{https://github.com/chi0tzp/UncPropCNN}{https://github.com/chi0tzp/uacnn}.
\end{abstract}

\section{Introduction}\label{sec:intro}
    
    In this technical report we study the problem of uncertainty propagation in typical building blocks of a Convolutional Neural Network (CNN), and give exact analytical solutions, or reasonable approximations, of the output of each operation. Input uncertainty is modeled in an element-wise fashion as structures of independent uni-variate Gaussian random variables with given means and variances. The output is computed in the same way, i.e., as structures of uni-variate Gaussians with means and variances given with respect to the corresponding input moments. For instance, in the case of a 2D convolution layer, its input is typically a 4D tensor, and we assume that each element of this tensor is an independent uni-variate Gaussian with given mean and variance. The output, again a 4D tensor, has elements that are independent uni-variate Gaussians.
    
\begin{figure}[t!]
        \centering
        \includegraphics[width=\columnwidth]{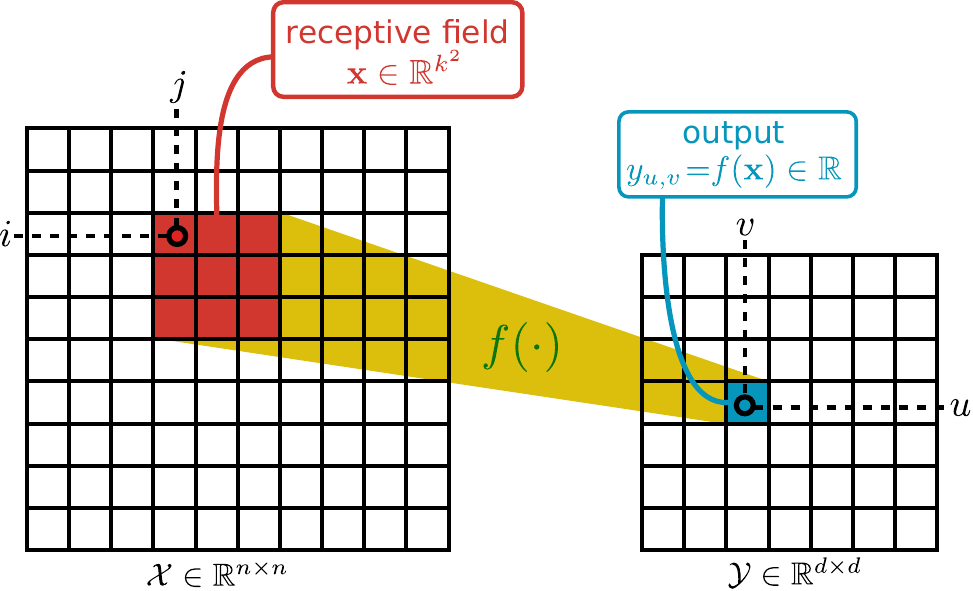}
        \caption{Basic filtering operation: at each step, the $k\times k$ filter $f$ acts on a receptive field $\mathbf{x}$ and produces an output $y=f(\mathbf{x})$.}
        \label{fig:filtering}
    \end{figure}

    We begin by introducing the terminology of the basic filtering operation, based on which many operations, such as convolution, are performed. Let $\mathcal{X}\in\mathbb{R}^{n\times n}$ be an input map that, after filtering, results in an output map $\mathcal{Y}\in\mathbb{R}^{d\times d}$. If $k$, $s$, and $p$ respectively denote the kernel, stride, and padding sizes of the filtering operation, then the dimension of the output map is given by $d=\floor*{\frac{n-k+2p}{s}}+1$. In Fig.~\ref{fig:filtering} we illustrate the operation of a $k\times k$ filter $f$. At each step, the filter acts on a $k\times k$ receptive field $\mathbf{x}$ and produces a scalar output $y=f(\mathbf{x})$. Filter operator $f$ can be either parametrized by learnable parameters (e.g., weights and biases in the case of a 2D convolution, i.e., $y=f(\mathbf{x};\mathbf{w},b)=\mathbf{w}^\top\mathbf{x}+b$), or perform a non-parametric operation, such as in the case of an average pooling operation.
    
    More specifically, by letting the above maps, i.e., $\mathcal{X}$ and $\mathcal{Y}$, be given as follows
    \begin{equation}\label{eq:X_matrix}
        \mathcal{X} = 
        \begin{pmatrix}
            x_{1,1} & x_{1,2} & \cdots & x_{1,n} \\
            x_{2,1} & x_{2,2} & \cdots & x_{2,n} \\
            \vdots  & \vdots  & \ddots & \vdots  \\
            x_{n,1} & x_{n,2} & \cdots & x_{n,n} 
        \end{pmatrix}
        \in\mathbb{R}^{n\times n},
    \end{equation}
    and
    \begin{equation}\label{eq:Y_matrix}
        \mathcal{Y} = 
        \begin{pmatrix}
            y_{1,1} & y_{1,2} & \cdots & y_{1,d} \\
            y_{2,1} & y_{2,2} & \cdots & y_{2,d} \\
            \vdots  & \vdots  & \ddots & \vdots  \\
            y_{d,1} & y_{d,2} & \cdots & y_{d,d} 
        \end{pmatrix}
        \in\mathbb{R}^{d\times d},
    \end{equation}
    we note that the $(u,v)$-th element of the output map $\mathcal{Y}$ is given by $y_{u,v} = f(\mathbf{x})$, where $\mathbf{x}$ is the $k^2$-dimensional vector representation of the following receptive field
    \begin{equation}\label{eq:receptive_field_matrix_X}
        \begin{pmatrix}
            x_{i,j}     & x_{i,j+1}     & \cdots & x_{i,j+k-1}      \\
            x_{i+1,j}   & x_{i+1,j+1}   & \cdots & x_{i+1,j+k-1}    \\
            \vdots      & \vdots        & \ddots & \vdots           \\
            x_{i+k-1,j} & x_{i+k-1,j+1} & \cdots & x_{i+k-1,j+k-1} 
        \end{pmatrix}
        \in\mathbb{R}^{k\times k},
    \end{equation}
    for some $i,j\in\{1,\ldots,n\}$ (see Fig.~\ref{fig:filtering}), and can be written as
    \begin{equation}\label{eq:receptive_field_vector_x}
        \mathbf{x}=\left(x_1,x_2,\ldots,x_{k^2}\right)^\top\in\mathbb{R}^{k^2}.
    \end{equation}
    As discussed above, we assume that each input element (thus, each element of an arbitrary receptive field) is a uni-variate Gaussian with given mean and variance, i.e., 
    $$
        x_{i}\sim\mathcal{N}\left(\mu_{x_i},\sigma^2_{x_i}\right),\:i=1,\ldots,k^2.
    $$
    As a result, each element of the output, $y_{u,v} = f(\mathbf{x})$ is also a random variable (Gaussian or otherwise) with an expected value given by $\mathbb{E}[y_{u,v}]=\mathbb{E}[f(\mathbf{x})]$ and a variance given by $\mathbb{V}[y_{u,v}]=\mathbb{V}[f(\mathbf{x})]$.
    
    The nature of the output random variable $y_{u,v}=f(\mathbf{x})$ (whether it preserves its normality or not), as well as whether it can be calculated analytically or be approximated, depends heavily on the filtering operation $f$. In the case of a linear operator, such as a standard 2D convolution, output moments can be computed in exact closed-form expressions, while in a non-linear one this is usually not tractable and, thus, appropriate approximation should be considered.

\section{Linear layers}\label{sec:linear}

    \subsection{Average pooling layer}\label{subsec:avgp}

\begin{figure}[t!]
            \centering
            \includegraphics[width=\columnwidth]{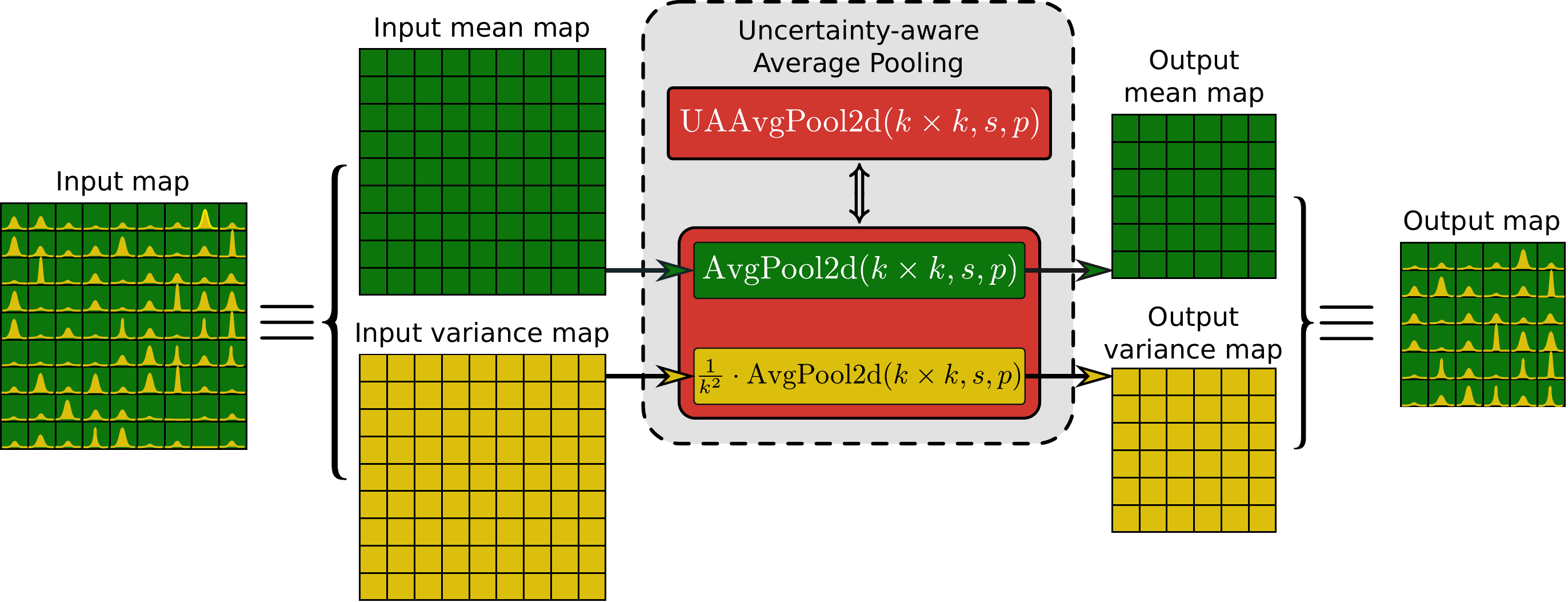}
            \caption{Uncertainty-aware average pooling (\texttt{UAAvgPool2d}).}
            \label{fig:uaavgp}
        \end{figure}

        Following the discussion above, average pooling can be seen as a non-parametric filtering operation, where $f$ simply computes the mean value of each receptive field; i.e., 
        $$
            f(\mathbf{x})=\frac{1}{k^2}\sum_{i=1}^{k^2}x_i.
        $$
        Then, the expected value and the variance of $f$ evaluated on an arbitrary receptive field $\mathbf{x}$ are respectively given as
        $$
            \mathbb{E}\left[f(\mathbf{x})\right]
            =
            \frac{1}{k^2}\sum_{i=1}^{k^2}\mathbb{E}[x_i]
            =
            \frac{1}{k^2}\sum_{i=1}^{k^2}\mu_{x_i}
        $$
        and
        $$
            \mathbb{V}\left[f(\mathbf{x})\right]
            =
            \frac{1}{k^2}\sum_{i=1}^{k^2}\mathbb{V}[x_i]
            =
            \frac{1}{k^4}\sum_{i=1}^{k^2}\sigma^2_{x_i}.
        $$
        Thus, the uncertainty-aware average pooling layer (\texttt{UAAvgPool2d}) can be performed using two standard average pooling layers; that is, one on the mean input map and one on the variance input map as shown in Fig.~\ref{fig:uaavgp}. The latter should be scaled by a factor of $\frac{1}{k^2}$.
        
    \subsection{Convolution layer}\label{subsec:conv}
        
        Adopting the terminology of Sect.~\ref{sec:intro}, we proceed to the calculation of the output of a 2D convolution operation given that its input is modeled as a set of uni-variate Gaussians with given moments. 
        
        As Fig.~\ref{fig:filtering} shown, the $(u,v)$-th element of the output map (for some $i,j\in\{1,\ldots,n\}$) is given by $y_{u,v}=f(\mathbf{x})$, which in this case is given by
        \begin{equation}\label{eq:y_u_v}
            y_{u,v} = \mathbf{w}^\top\mathbf{x}+b,
        \end{equation}
        where $\mathbf{w}$ denotes the row-wise, $k^2$-dimensional vector representation of the $k\times k$ weight matrix
        \begin{equation}\label{eq:weight_matrix_W}
            W = 
            \begin{pmatrix}
                w_{1,1} & w_{1,2} & \cdots & w_{1,k}  \\
                w_{2,1} & w_{2,2} & \cdots & w_{2,k}  \\
                \vdots  & \vdots  & \ddots & \vdots   \\
                w_{k,1} & w_{k,2} & \cdots & w_{k,k} 
            \end{pmatrix}
            \in\mathbb{R}^{k\times k},
        \end{equation}
        i.e.,
        \begin{equation}\label{eq:weight_vector_w}
            \mathbf{w}=\left(w_1,w_2,\ldots,w_{k^2}\right)^\top\in\mathbb{R}^{k^2},
        \end{equation}
        $b$ the convolution's bias term, and $\mathbf{x}$ the row-wise, $k^2$-dimensional vector representation of the receptive field.
        
        \begin{figure}[t!]
            \centering
            \includegraphics[width=\columnwidth]{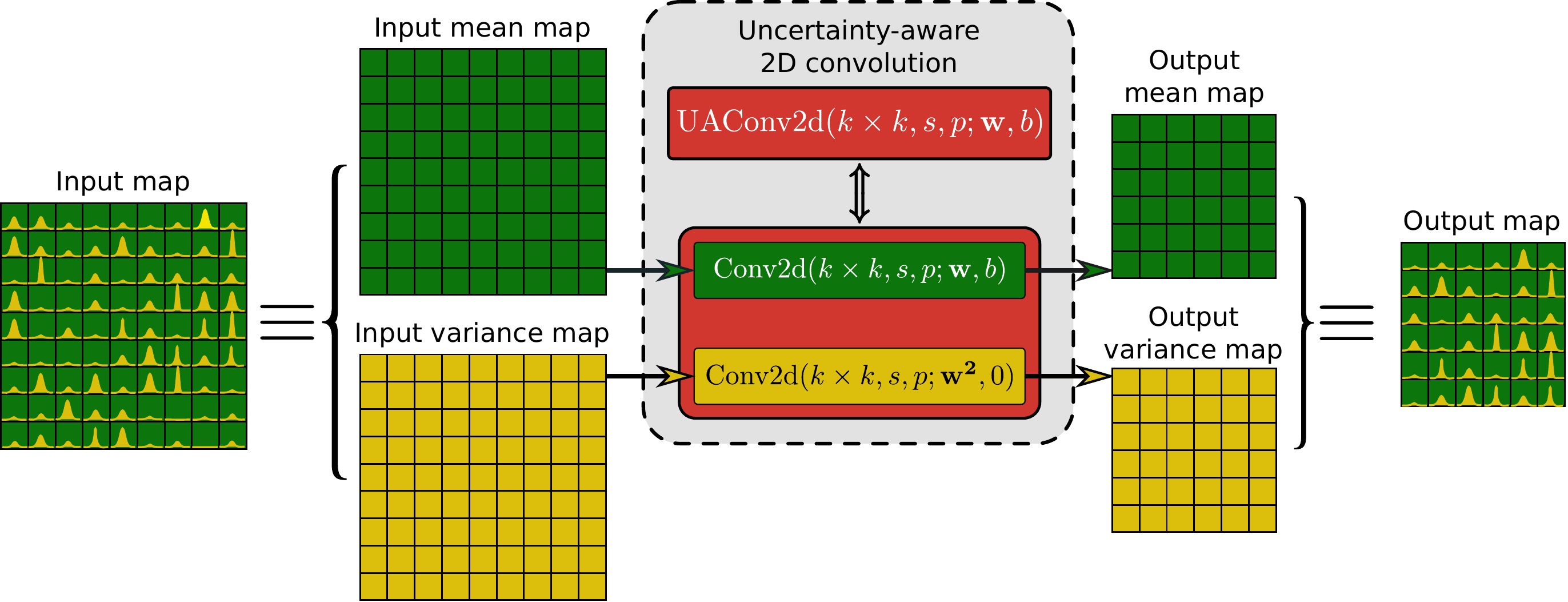}
            \caption{Uncertainty-aware $2$D convolutional operation (\texttt{UAConv2d}).}
            \label{fig:uaconv2d}
        \end{figure}
    
        Let us now assume that each element of the input map $\mathcal{X}$, i.e, each $x_{i,j}\in\mathbb{R}$, $i,j\in\{1,\ldots,n\}$, is a uni-variate Gaussian variable with given mean and variance. That is,
        \begin{equation}\label{eq:x_ij}
            x_{i,j}\sim\mathcal{N}\left(\mu_{x_{i,j}},\sigma^2_{x_{i,j}}\right),\:i,j\in\{1,\ldots,n\}.
        \end{equation}
        As a consequence, the vector representation of the receptive field $\mathbf{x}$ is a multivariate Gaussian vector with given mean 
        \begin{equation}\label{eq:mu_x}
            \bm{\mu}_x = \left(\mu_{x_{i,j}},\mu_{x_{i,j+1}},\ldots,\mu_{x_{i+k-1,j+k-1}}\right)^\top\in\mathbb{R}^{k^2}
        \end{equation}
        and (diagonal) covariance matrix 
        \begin{equation}\label{eq:Sigma_x}
            \Sigma_x=
            \operatorname{diag}\left(\sigma^2_{x_{i,j}},\sigma^2_{x_{i,j+1}},\ldots,\sigma^2_{x_{i+k-1,j+k-1}}\right)\in\mathbb{S}^{k^2}_{++}.
        \end{equation}
        That is, $\mathbf{x}\sim\mathcal{N}\left(\bm{\mu}_x,\Sigma_x\right)$. For the sake of convenience, we represent the above diagonal covariance matrix as a $k^2$-dimensional vector
        \begin{equation}\label{eq:sigma_2_x}
            \bm{\sigma}^{\mathbf{2}}_x=\left(\sigma^2_{x,1},\sigma^2_{x,2},\ldots,\sigma^2_{x,k^2}\right)^\top\in\mathbb{R}^{k^2}. 
        \end{equation}
        Thus, due to (\ref{eq:y_u_v}), $y_{u,v}\sim\mathcal{N}\left(\mu_{y_{u,v}},\sigma^2_{y_{u,v}}\right)$, where
        \begin{equation}\label{eq:mu_y_u_v}
            \mu_{y_{u,v}} = \mathbf{w}^\top\bm{\mu}_x+b,
        \end{equation}
        and
        \begin{equation}\label{eq:sigma2_y_u_v}
            \sigma^2_{y_{u,v}} = \mathbf{w}^\top\Sigma_x\mathbf{w}.
        \end{equation}
        This means that the $(u,v)$-th element of the output map $\mathcal{Y}$ is itself a uni-variate Gaussian variable, whose first- and second-order moments are given with respect to the moments of the corresponding receptive field (see Fig.~\ref{fig:uaconv2d}). Furthermore, since a Gaussian distribution is uniquely characterized by its first- and second-order moments, we represent input map $\mathcal{X}$ with a pair of maps, one for the mean values of $x_{i,j}$'s and one for their variances (see leftmost side of Fig.~\ref{fig:uaconv2d}). Let them be $M_{\mathcal{X}}$ and $V_{\mathcal{X}}$, respectively. Likewise, it suffices to have a pair of maps for the output $\mathcal{Y}$, the means map $M_{\mathcal{Y}}$ and variances map $V_{\mathcal{Y}}$ (see rightmost side of Fig.~\ref{fig:uaconv2d}).
        
        \begin{figure}[t!]
            \centering
            \includegraphics[width=\columnwidth]{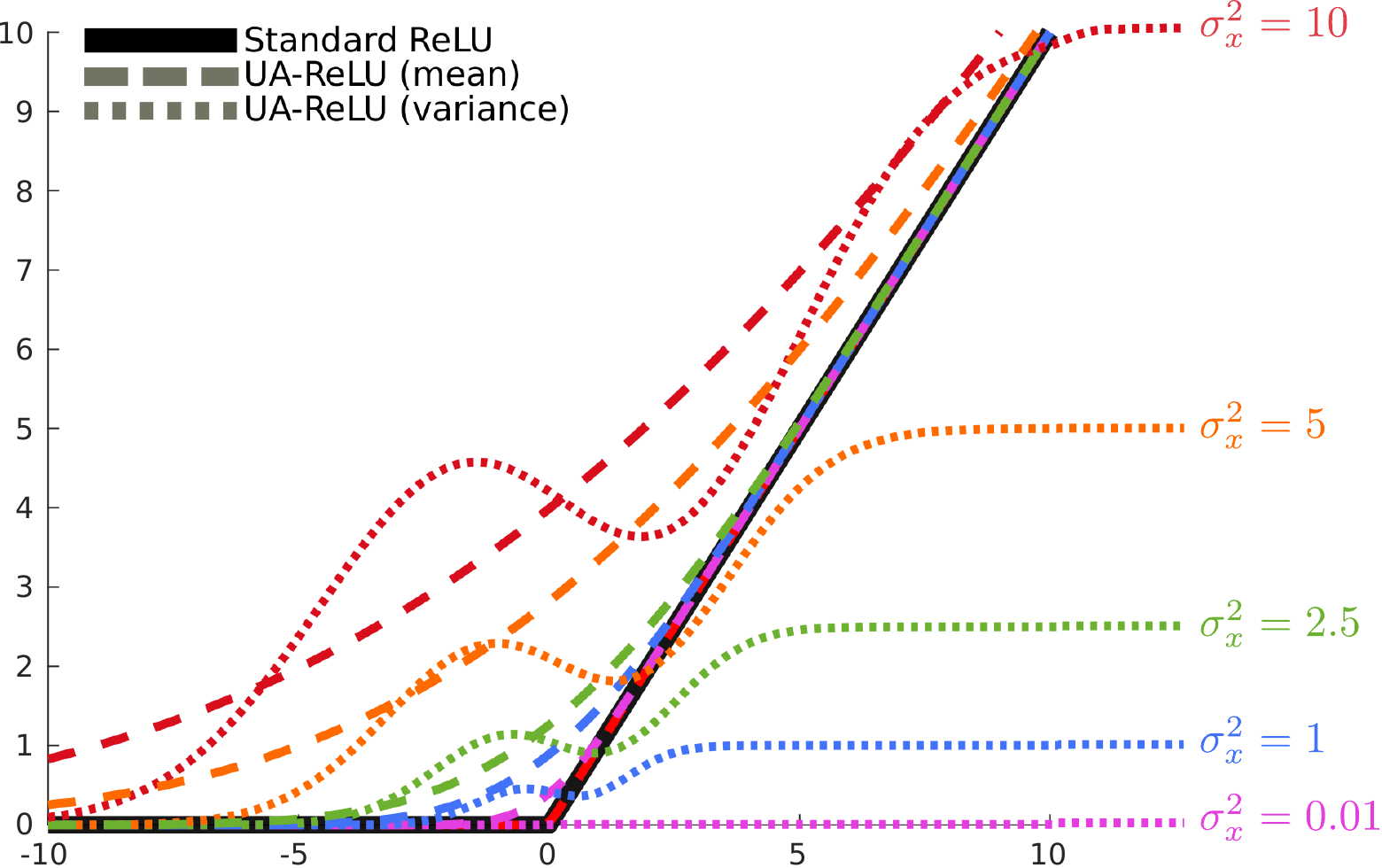}
            \caption{Uncertainty-aware Rectified Linear Unit (\texttt{UAReLU}) for various amounts of input uncertainty.}
            \label{fig:uarelu_mean_var_plot}
        \end{figure}
        
        Following the discussion above, it can easily be verified that the means map of the output, $M_{\mathcal{Y}}$, can be obtained by applying a $2$D convolution (kernel size $k$, stride $s$, and padding $p$) on the means map of the input, $M_{\mathcal{X}}$. 
                
        In the case of the variances map of the output, let us revisit (\ref{eq:sigma2_y_u_v}). Since the covariance matrix $\Sigma_x$ is diagonal (see (\ref{eq:Sigma_x})), the quadratic form of (\ref{eq:sigma2_y_u_v}) boils down to an inner product, that is
        \begin{equation}\label{eq:sigma2_y_u_v_inner_product}
            \sigma^2_{y_{u,v}} = \mathbf{w}^\top\Sigma_x\mathbf{w}=\sum_{t=1}^{k^2}w_t^2\sigma^2_{x,t}=\mathbf{w^2}^\top\bm{\sigma}^{\mathbf{2}}_x,
        \end{equation}
        where $\mathbf{w^2}$ merely denotes the element-wise square of the weights vector $\mathbf{w}$, i.e., 
        \begin{equation}\label{eq:weight_vector_w_squared}
            \mathbf{w^2}=\left(w^2_1,w^2_2,\ldots,w^2_{k^2}\right)^\top\in\mathbb{R}^{k^2},
        \end{equation}
        and $\bm{\sigma}_x^{\mathbf{2}}$ is the vector of variances of the elements of the corresponding receptive field (given in (\ref{eq:sigma_2_x})).
                
        It can easily be verified, thus, that the variance map of the output, $V_{\mathcal{Y}}$, can be obtained by applying a $2$D convolution (kernel size $k$, stride $s$, and padding $p$) on the variance map of the input, $V_{\mathcal{X}}$, with zero bias ($b=0$) and weights the element-wise square of the weights used in the previous convolution (the one applied on the input means map in order to produce the output means map).
    
        The above are illustrated in Fig.~\ref{fig:uaconv2d}, where the uncertainty-aware $2$D convolution (\texttt{UAConv2d}) is defined as a pair of standard 2D convolutions. More specifically, one 2D convolution operation (parametrized by a set of weights $\mathbf{w}$ and bias $b$) is used in order to produce the mean values of the output map, while a second $2$D convolution operation (parametrized by the element-wise squares of the \textit{same} weights $\mathbf{w}$ and zero bias) is used in order to produce the variances of the output map. Hence, \texttt{UAConv2d} does not increase the number of learnable parameters compared to a standard 2D convolution. 
    
    \subsection{Linear (fully-connected) layer}\label{subsec:fc}
\begin{figure}[t!]
            \centering
            \includegraphics[width=\columnwidth]{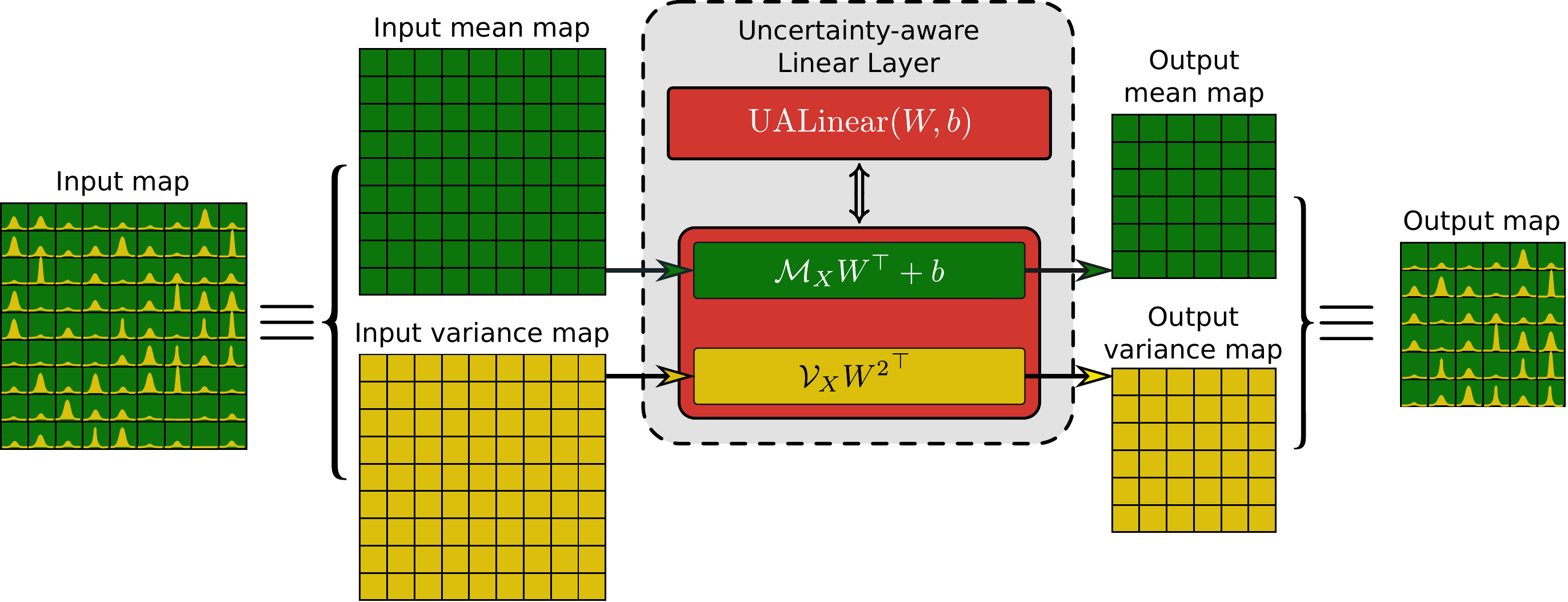}
            \caption{Uncertainty-aware linear layer (\texttt{UALinear}).}
            \label{fig:uafc}
        \end{figure}

        A fully-connected layer performs the linear operation $\mathcal{Y}=\mathcal{X}W^\top+b$, where $W$ is the weights matrix and $b$ the biases vector. Following similar arguments as above, we compute the output mean and variance maps, respectively as 
        $$
        \mathcal{M}_Y = \mathcal{M}_X W^\top + b,\quad\text{and}\quad\mathcal{V}_Y = \mathcal{V}_X {W^2}^\top,
        $$
        where $W^2$ is the element-wise square of the parameters matrix $W$. This uncertainty-aware linear operation (\texttt{UALinear}) can be performed using two standard linear (fully-connected) layers, one for the mean input map ($\mathcal{M}_X$) and one for the variance input map ($\mathcal{V}_X$) as shown in Fig.~\ref{fig:uafc}.
    
\section{Non-linear layers}\label{sec:nonlinear}

    
    \subsection{Rectified Linear Unit (ReLU)}\label{subsec:relu}
    
    
        Rectified Linear Unit (ReLU) is defined as $h\colon\mathbb{R}\to\mathbb{R}_+$:
        \begin{equation}\label{eq:relu}
            h(x) = \max(0,x)
        \end{equation}
        Let $x$ be a uni-variate normal variable with given mean and variance, i.e., $x\sim\mathcal{N}\left(\mu_x,\sigma^2\right)$ and probability density function $f\colon\mathbb{R}\to\mathbb{R}_+$ given as
        $$
            f(x) = \frac{1}{\sqrt{2\pi\sigma_x^2}}\exp\left(-\frac{(x-\mu_x)^2}{2\sigma_x^2}\right).
        $$
        Then, the expected value and the variance of ReLU are given, respectively, as follows:
        \begin{align}
            \mu_h = \mathbb{E}[h(x)] = & \int_{\mathbb{R}}h(x)f(x)\mathrm{d}x \nonumber \\
                    = & \int_{\mathbb{R}}\max(0,x)f(x)\mathrm{d}x=\int_0^{\infty}xf(x)\mathrm{d}x,
            \label{eq:relu_mean_def}
        \end{align}
        and
        \begin{align}
            \sigma^2_h = \mathbb{V}[h(x)] = & \mathbb{E}[h^2(x)]-\left(\mathbb{E}[h(x)]\right)^2 \nonumber \\
                       = & \int_0^{\infty}x^2f(x)\mathrm{d}x-\mu_h^2.
            \label{eq:relu_var_def}
        \end{align}
        By evaluating the above integrals, we arrive at the following closed-form expressions:
        \begin{equation}\label{eq:relu_mean_res}
            \mu_h=
            \sqrt{\frac{\sigma_x^2}{2\pi}}\exp\left(-\frac{\mu_x^2}{2\sigma_x^2}\right)
            +
            \frac{\mu_x}{2}
            \left(1+\operatorname{erf}\left(\frac{\mu_x}{\sqrt{2\sigma_x^2}}\right)\right),
        \end{equation}
        and
        \begin{align}
            \sigma^2_h = & \frac{\sigma_x^2+\mu_x^2}{2}\left(1+\operatorname{erf}\left(\frac{\mu_x}{\sqrt{2\sigma_x^2}}\right)\right)
            - \frac{\mu_x\sigma_x}{\sqrt{2\pi}}
            \exp\left(-\frac{\mu_x^2}{2\sigma_x^2}\right)\nonumber \\
                    & -\mu_h^2.
            \label{eq:relu_var_res}
        \end{align}
        Fig.~\ref{fig:uarelu_mean_var_plot} illustrates the expected value (dashed lines) and the variance (dotted lines) of the uncertainty-aware ReLU (\texttt{UAReLU}) for various amounts of input uncertainty (depicted in different colours).
    
    \subsection{Sigmoid function}\label{subsec:sigmoid}
        Let $s\colon\mathbb{R}\to(0,1)$ be the sigmoid function given as
        \begin{equation}
            s(x) = \frac{1}{1+\exp(-x)}.
        \end{equation}
                
        Also, let $x\sim\mathcal{N}(\mu,\sigma^2)$, be a uni-variate normal variable. Following similar arguments as in~\cite{sigmoid}, we approximate the expected value and the variance of $s(x)$ respectively as follows
        \begin{equation}\label{eq:sigmoid_mean}
            \mathbb{E}[s(x)] = s\left(\frac{\mu}{\sqrt{1+\lambda\sigma^2}}\right),
        \end{equation}
        and
        \begin{align}
            \mathbb{V}[s(x)] & =             s\left(\frac{\mu}{\sqrt{1+\frac{3\sigma^2}{\pi^2}}}\right)\cdot\left[1-s\left(\frac{\mu}{\sqrt{1+\frac{3\sigma^2}{\pi^2}}}\right)\right]\cdot \nonumber \\
            & \quad\quad\left(1-\frac{1}{\sqrt{1+\frac{3\sigma^2}{\pi^2}}}\right).
            \label{eq:sigmoid_var}
        \end{align}

\section{Loss functions}\label{sec:loss}

    \subsection{Binary cross-entropy loss function}\label{subsec:bceloss}
    
\begin{figure}[ht!]
            \centering
            \includegraphics[width=0.8\columnwidth]{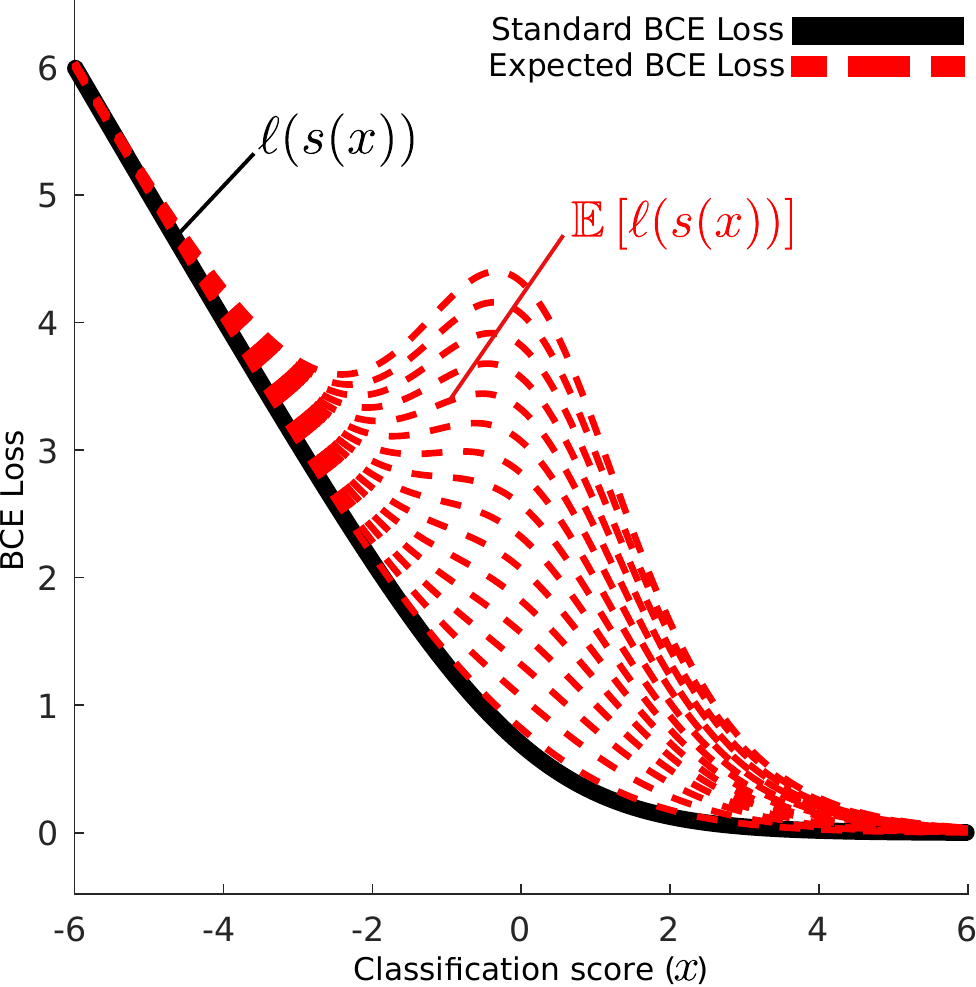}
            \caption{Expected BCE loss (\texttt{UABCELoss}) for various amounts of input uncertainty (dashed red lines) compared to standard BCE loss.}
            \label{fig:uabce_mean}
        \end{figure}

        Binary cross-entropy (BCE) loss introduced by a training sample with prediction score $s(x)$, i.e., after applying the simgoid function, and truth label $y\in\{0,1\}$, is given by
        \begin{equation}\label{eq:bce_loss_1}
            \ell(x) = -\left[y\log(s(x))+(1-y)\log(1-s(x))\right].
        \end{equation}
        Using that $\log(1-s(x))=\log(s(x))-x$, the above can be rewritten as follows
        \begin{equation}\label{eq:bce_loss_2}
            \ell(x) = -\log(s(x)) + x(1-y).
        \end{equation}
        Thus, under the assumption of random input $x\sim\mathcal{N}(\mu_x,\sigma_x^2)$, the expected BCE loss is given as
        \begin{align}
            \mathbb{E}[\ell(x)] = & -\mathbb{E}[\log(s(x))] - (1-y)\mathbb{E}[x] = -\mathbb{E}[\log(s(x))] \nonumber \\
            & -(1-y)\mu_x.
        \end{align}
        We approximate the expectation of the log-sigmoid mapping using a second-order Taylor expansion. That is, 
        \begin{equation}
            \frac{\mathrm d}{\mathrm dx}\log\left(s(x)\right)=1-\frac{\exp(x)}{1+\exp(x)} = 1-s(x)
        \end{equation}
        and
        \begin{equation}
            \frac{\mathrm d^2}{\mathrm dx^2}\log\left(s(x)\right)=s(x)\left(s(x)-1\right).
        \end{equation}
        We note that the log-sigmoid mapping $\log\left(s(x)\right)$ is related to the anti-derivative of the sigmoid mapping $s(x)$. Using a second-order Taylor approximation, the expectation of the log-sigmoid mapping is given as
        \begin{equation}
            \mathbb{E}\left[\log\left(s(x)\right)\right]
            \approx
            \log\left(s(\mu_x)\right)-\frac{1}{2}s(\mu_x)\left(1-s(\mu_x)\right)\sigma_x^2.
        \end{equation}
        Thus, the expected BCE loss can be approximated as follows
        \begin{equation}\label{eq:bce_loss_mean}
            \mathbb{E}[\ell(x)]=-\log\left(s(\mu_x)\right)-\frac{1}{2}s(\mu_x)\left(1-s(\mu_x)\right)\sigma_x^2-(1-y)\mu_x.
        \end{equation}
        Fig.~\ref{fig:uabce_mean} illustrates the expected value of BCE loss (\texttt{UABCELoss} -- red dashed lines, for various values of $\sigma_x$) compared to the standard BCE loss (black solid line).
    
\bibliographystyle{ieeetr}
\bibliography{references}

\end{document}